\documentclass{article}
\usepackage{ICASSP2021/spconf,amsmath,graphicx}
\usepackage{hyperref}
\usepackage[dvipsnames]{xcolor}
\usepackage{amssymb}
\usepackage{array,multirow}
\usepackage{float}
\title{Image coding for machines: an end-to-end learned approach}
\name{\begin{tabular}{c}Nam Le$^{\ast}$, Honglei Zhang$^{\dag}$, Francesco Cricri$^{\dag}$,  
    Ramin Ghaznavi-Youvalari$^{\dag}$, Esa Rahtu$^{\ast}$\end{tabular}}
\address{$^{\dag}$Nokia Technologies, $^{\ast}$Tampere University \\ 
Tampere, Finland}

\begin{document}
%
\maketitle


\newcommand{\aefigwidth}{1.0}
\newcommand{\outputfigwidth}{0.495}

\newcommand{\newparagraph}[1]{\par\textbf{#1}}

\newcommand{\Tensor}[1]{\boldsymbol{#1}}
\newcommand{\Loss}[1]{\mathcal{L}_{#1}}
\newcommand{\Model}[2]{\boldsymbol{#1}({#2})}
\newcommand{\ModelWeights}[1]{\boldsymbol{\theta}_{\boldsymbol{#1}}}
\newcommand{\Wmodel}[2]{\Model{#1}{{#2};\ModelWeights{#1}}}

\newcommand{\img}{\Tensor{x}}
\newcommand{\resimg}{\Tensor{\hat{x}}}
\newcommand{\latent}{\Tensor{y}}
\newcommand{\iqlatent}{\Tensor{\hat{y}}}
\newcommand{\tqlatent}{\Tensor{\Tilde{y}}}
\newcommand{\hyperlatent}{\Tensor{z}}
\newcommand{\tqhyperlatent}{\Tensor{\Tilde{z}}}
\newcommand{\iqhyperlatent}{\Tensor{\hat{z}}}
\newcommand{\prior}[1]{p_{#1}}
\newcommand{\encoder}[1]{\Wmodel{E}{#1}}
\newcommand{\decoder}[1]{\Wmodel{D}{#1}}
\newcommand{\probmodel}[1]{\Wmodel{P}{#1}}
\newcommand{\wrate}{w_{rate}}
\newcommand{\wtask}{w_{task}}
\newcommand{\wmse}{w_{mse}}
\newcommand{\wset}{\mathcal{W}}
\newcommand{\lossrate}{\Loss{rate}}
\newcommand{\losstask}{\Loss{task}}
\newcommand{\lossmse}{\Loss{mse}}

\newcommand{\BDrateDet}{37.87}
\newcommand{\BDrateSeg}{32.90}
\begin{abstract}
Over recent years, deep learning-based computer vision systems have been applied to images at an ever-increasing
pace, oftentimes representing the only type of consumption for those images. Given the dramatic explosion in 
the number of images generated per day, a question arises: how much better would an image codec targeting 
machine-consumption perform against state-of-the-art codecs targeting human-consumption? In this paper, 
we propose an image codec for machines which is neural network (NN) based and end-to-end learned. In 
particular, we propose a set of training strategies that address the delicate problem of balancing competing 
loss functions, such as computer vision task losses, image distortion losses, and rate loss. Our experimental 
results show that our NN-based codec outperforms the state-of-the-art Versatile Video Coding (VVC) standard 
on the object detection and instance segmentation tasks, achieving -\BDrateDet\% and -\BDrateSeg\% of BD-rate 
gain, respectively, while being fast thanks to its compact size.
    \ifdefined\ofincluded
        In addition, we propose a novel 
        inference-time optimization of the latent tensor which further boosts 
        the rate-distortion performance of the pretrained codec.
    \fi
To the best of our knowledge, this is the first end-to-end learned machine-targeted image codec.
\end{abstract}
\section{Introduction and background}
\label{sec:intro}
Over the years, traditional image and video coding standards such as Versatile 
Video Coding (VVC) \cite{vvc} have significantly 
improved the coding efficiency or rate-distortion performance, in which the ``distortion'' 
is measured by metrics aimed at improving the quality for human consumption. 
In the era of deep learning, computer vision tasks such as object detection or 
image classification represent a significant portion of image and video consumers, 
oftentimes being the only ones, e.g. self-steering system for cars in \cite{selfdriving}.
In this paper, we refer to computer vision tasks simply as \textit{machines}.
These \textit{machines} are usually applied (and trained) on human-targeted compressed 
images, which contain distortions that are less perceivable by humans but may 
lower the performance of the computer vision tasks. 
In addition, a lot of information contained in these compressed images is not 
necessary for a neural network (NN) to perform a task, such as detecting objects. 
Thus, it is likely that higher coding efficiency can be achieved by designing a 
codec with the specific goal of targeting \textit{machines} as the only consumer. 

In order to directly improve the task performance, \cite{vcm_feature_based}
and \cite{adapting_JPEGXS} propose standard-compliant methods that preserve the 
standard coded bitstream format by fine-tuning specific parts of the traditional codec
for the targeted \textit{machines}. Although the above methods manage to improve the task performance, 
they do not aim to completely replace the conventional pipeline for human-oriented coding, 
instead they only add an incremental capability to the system.
The use of neural networks to aid or completely replace the traditional 
image codec for human consumption has been actively studied recently.
In \cite{jointlearnedvvc}, a CNN post-processing filter is used to 
enhance the image quality. Some other proposals seek to effectively 
model the distribution of the data that is encoded/decoded by a lossless codec. 
To this end, \cite{pixelCNN} proposes an autoregressive scheme for distribution modeling, 
while the system in \cite{l3c} hierarchically learns the distribution through feature maps
at multiple scales.
In \cite{scale_hyperprior, meanscale_hyperprior}, the authors 
propose end-to-end pipelines for image \textit{transform coding}, 
where the image's latent representation and its distribution are entropy encoded 
using hierarchical learned \textit{hyperprior}.
Our work can be viewed as an extension of \cite{meanscale_hyperprior}, in which 
the main targeted end-users are \textit{machines}, and the main coder is 
substituted by a more capable one.
\ifdefined\ofincluded
    In \cite{weightupdate_paper}, inference-time 
    fine-tuning schemes is proposed, aimed at obtaining better compression rates, 
    while \cite{jpegai_paper} extensively studies latent tensor overfitting.
    However, as these methods target human consumption, they are only useful for 
    enhancing the fidelity in pixel domain, where ground-truth at encoder-side is 
    easily available (the uncompressed image itself) and no manually-generated 
    labels are required.
\fi
Regarding the problem of multi-objective training, \cite{lossweighting_overview} 
discusses the importance of loss weighting and methods for dynamic loss balancing, 
as opposed to fixed weighted losses.
These methods, however, do not directly offer full control over the 
priorities of the objectives, which is a desired feature in our case.

In this paper, we propose an end-to-end learned system that directly optimizes 
the rate-distortion trade-off, where the distortion is the training loss of the pretrained 
task-NN. 
In order to achieve better coding efficiency than state-of-the-art traditional 
codecs, we introduce an adaptive loss weighting strategy addressing the 
problem of balancing competing losses in multi-task learning\ifdefined\ofincluded
    , along with 
    a novel inference-stage optimization technique for the latent tensor output by 
    the encoder
\fi.
In our experimental section, we test our proposed techniques on the tasks of 
object detection and instance segmentation, and show that our system outperforms 
VVC significantly.
\section{Proposed method}
\label{sec:proposed}
\begin{figure}
    \begin{minipage}[b]{1.0\linewidth}
      \centering
      \centerline{\includegraphics[trim=0cm 0cm 1.0cm 0cm, width=9.0cm]
      {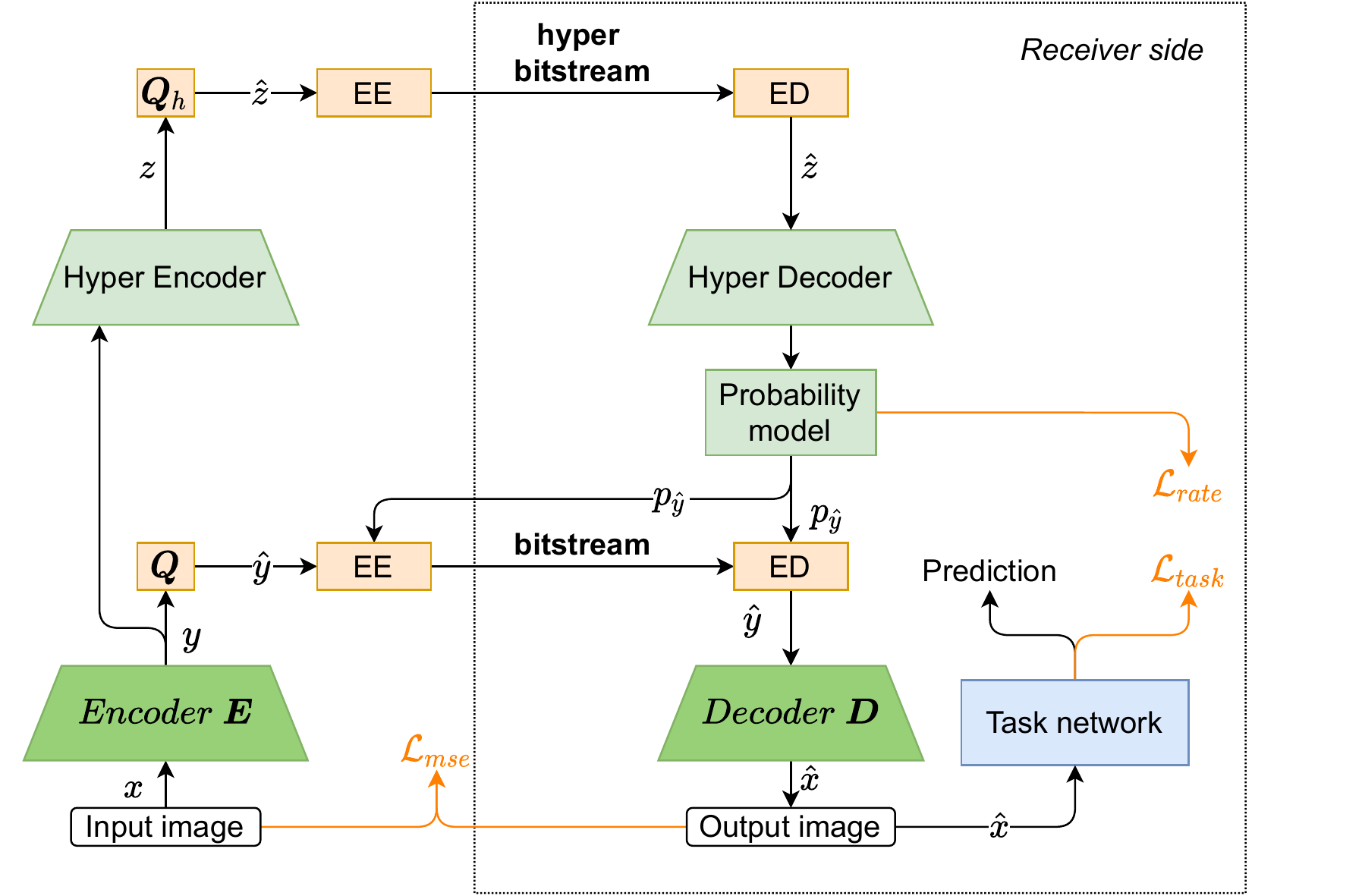}}
    \end{minipage}
  \caption[Pipeline overview]{Image Coding for Machines overview. 
  ``EE'' and ``ED'' denote entropy encoders and decoders, respectively.}
  \label{fig:overview}
\end{figure}

In contrast to the pipeline in \cite{meanscale_hyperprior}, we 
propose an Image Coding for Machines (ICM) system targeting task performance 
instead of pixel-domain fidelity.
Our codec comprises a neural auto-encoder, a learned probability model and an 
entropy codec.
The proposed pipeline is illustrated in \autoref{fig:overview}.
The NN-based encoder transforms the uncompressed image 
$\img$ to a new data representation $\latent = \encoder{\img}$, which is then 
quantized as $\iqlatent = \Model{Q}{\latent}$ and subsequently 
lossless-compressed by an entropy encoder, 
using the probability distribution estimated by the probability model. The 
output bitstream is decompressed on the decoder-side and decoded back to the pixel domain by the NN decoder as 
$\resimg = \decoder{\iqlatent}$. 
The task NN takes $\resimg$ as input and returns the corresponding 
task results.
\subsection{Auto-encoder}
\label{sec:autoencoder}
\begin{figure}[ht]
    \begin{minipage}[b]{\linewidth}
      \centering
      \centerline{\includegraphics[trim=0cm 0cm 1.0cm 0cm, 
      width=\aefigwidth\linewidth]{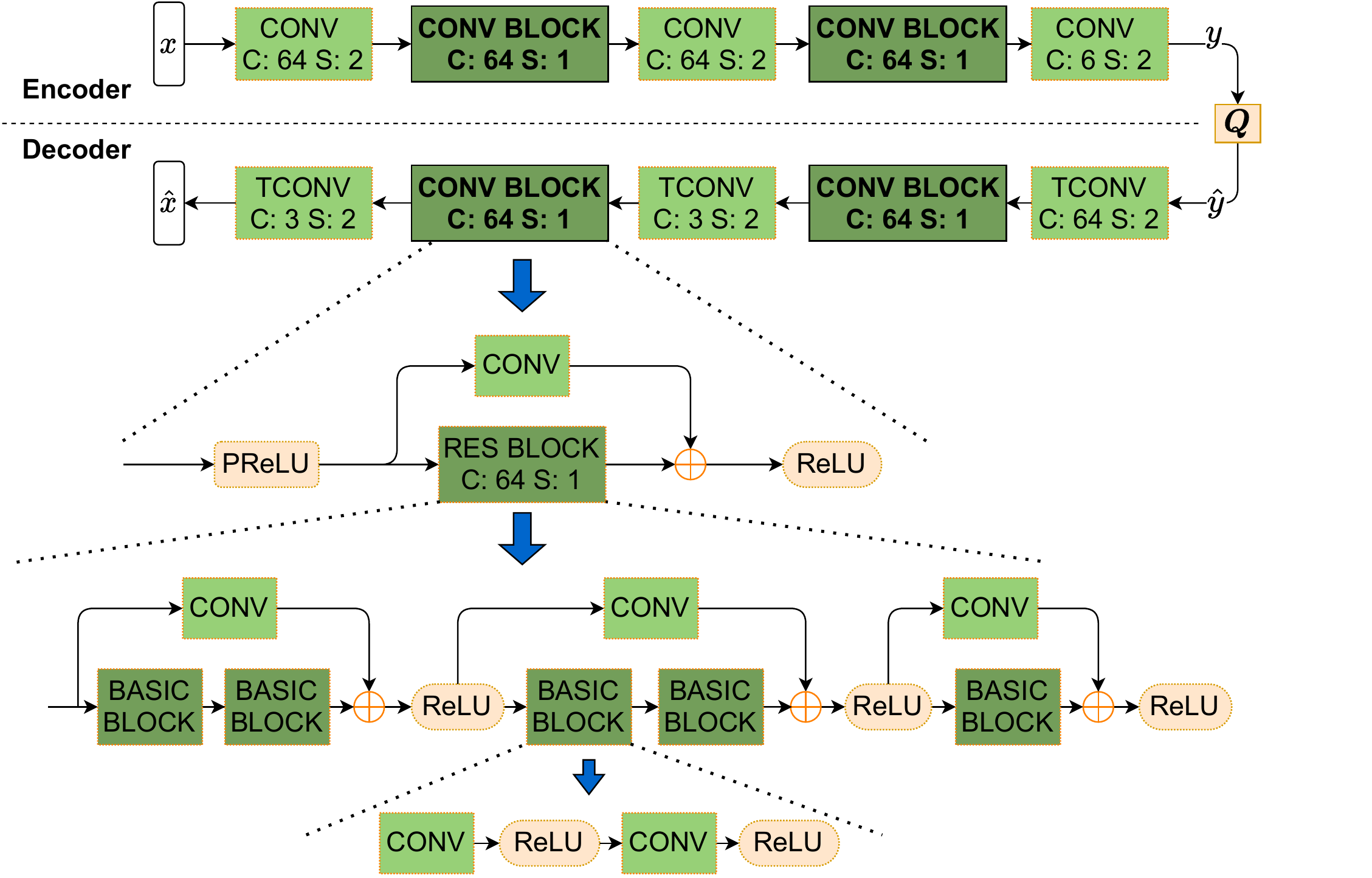}}
    \end{minipage}
\caption{Auto-encoder architecture. The convolutional blocks are illustrated by 
sharp rectangles. ``TCONV'' denotes the transposed convolutional layers. In each 
convolutional block, ``S'' denotes the stride and ``C'' denotes number of output 
channels for all of the children blocks. These values are inherited from the 
parent block if not stated otherwise.}
\label{fig:autoencoder_arch}
\end{figure}
Unlike common auto-encoders, our proposed auto-encoder does not aim to reconstruct the 
input image. Instead, its goal is to decode a data tensor that can provide a 
good task performance, while the encoder's output can be efficiently compressed 
by the entropy codec. 
These two objectives are referred to as task loss $\losstask$ and rate loss 
$\lossrate$, respectively. 
As the task NNs are already pretrained and left unmodified, they accept input data in the format of 
images, i.e., three channels. 
Thus, the output of the decoder needs to be a tensor of same shape 
as an image. 
For the architecture of the encoder and decoder, we use a convolutional 
neural network (CNN) architecture with residual connections, 
as illustrated in \autoref{fig:autoencoder_arch}. 
In order to keep encoding-decoding time and resource consumption low, we chose 
to use CNNs with a small number of filters in the intermediate and last layers. 
This auto-encoder is optimized by using the aforementioned loss terms.
\subsection{Probability model}
\label{ssec:probmodel}
An asymmetric numeral systems (ANS) \cite{ans} codec first encodes a stream of 
symbols into a single natural number according to the probability of each 
symbol, then it converts the number into a binary bitstream. 
Given the quantized latent tensor $\iqlatent$ and its estimated distribution 
$\prior{\iqlatent}(\iqlatent)$, we're interested to encode the symbols of 
$\iqlatent$ with a minimum code length lower-bounded by Shannon entropy. 
The lower-bound is achieved if the marginal distribution $m_{\iqlatent}$ is 
identical to $\prior{\iqlatent}$. 
Since $m_{\iqlatent}$ arises from the unknown input image distribution 
$\prior{\img}$ and the transformation method $\Model{Q}{\encoder{\cdot}}$, 
the code length $r$ can only be estimated by Shannon cross-entropy:
\begin{equation}
    \begin{split}
        r
        & = \mathbb{E}_{\iqlatent \sim m_{\iqlatent}} \left[-\log_{2} \prior{
            \iqlatent}(\iqlatent)\right] \\
        & = \mathbb{E}_{\img \sim \prior{\img}}\left[-\log_{2} \prior{\iqlatent}
        (\Model{Q}{\encoder{\img}})\right] \\
    \end{split}
\end{equation}
The probability model aims to learn the distribution $\prior{\iqlatent}$ in 
order to minimize $r$. 
For this module, we use the ``Mean \& Scale Hyperprior'' structure proposed in 
\cite{meanscale_hyperprior} which models the latent by Gaussian mixture models 
and learns the parameters to these distributions. 
The distribution $\prior{\iqlatent}$ is obtained on-the-fly using these 
parameters in a closed-form fashion. 
In order to decompress the coded latents $\iqlatent$, an additional 
bitstream for the ``hyper-latents'' $\iqhyperlatent$ carrying the hyperprior embeddings
is sent to the receiver side. 
The rate loss term $\lossrate$ is thus given by the total length of the two 
bitstreams:
\begin{equation}
    \begin{split}
        \lossrate = \underbrace{\mathbb{E}_{\iqlatent \sim m_{\iqlatent}}\left[
            -\log_{2} \prior{\iqlatent}(\iqlatent)\right]}_{\text{latents rate}} 
            +  \underbrace{\mathbb{E}_{\iqhyperlatent \sim m_{\iqhyperlatent}}\left[-\log_{2} 
            \prior{\iqhyperlatent}(\iqhyperlatent)\right]}_{\text{hyper-latents 
            rate}}
    \end{split}
    \label{eq:rate_loss}
\end{equation}
The probability model is jointly optimized with the auto-encoder in an 
end-to-end training fashion. 
During training, the quantization step is replaced by additive uniform noise to make 
gradient-based optimization possible \cite{scale_hyperprior,relaxed_quantization}.

\subsection{Training strategy}
\label{ssec:strategy}
We separately trained and evaluated two different compression models for two 
computer vision tasks: object detection, using Faster R-CNN \cite{fasterRCNN}, 
and instance segmentation, using Mask R-CNN \cite{maskRCNN}. 
In each case, we freeze the corresponding pre-trained task network and define 
$\losstask$ as the respective training task loss. 
Thus, gradients of $\losstask$ are computed only with respect to the codec's 
parameters.

Image coding is often posed as a rate-distortion optimization (RDO) problem, 
i.e.  $J = R + \lambda\cdot D$, where $R,D$ and $J$ denote the bitrate, the distortion
and joint cost, respectively, and $\lambda$ is the 
Lagrange multiplier driving the trade-off between them: more encoding bits 
(high $R$) reduces the distortion (low $D$) and vice versa \cite{vvc,l3c,
meanscale_hyperprior}. 
The common way of performing RDO when training NN-based codecs is 
to find a certain set of values for $\lambda$ (and other hyper-parameters) so that the 
desired rate-distortion is achieved 
after a number of training iterations \cite{l3c,meanscale_hyperprior}. 
The learned models are then saved as compression models to be used for the 
corresponding bitrates. 
Given that the decoded images are consumed by machines, it is not 
necessary to have high fidelity output images that are visually appealing to 
humans. 
We instead prioritize good task performance by imposing task loss $\losstask$ 
minimization on our model training.
Our ICM system extends the 
above RDO approach by adding the task loss $\losstask$ to the ``distortion'' $D$. 
The general training loss function is given by:
\begin{equation}
    \Loss{total} = \wrate\lossrate + \wmse\lossmse + \wtask\losstask ,
\end{equation}
where $\wrate, \wmse, \wtask$ are the scalar weights for each loss term 
$\lossrate, \lossmse, \losstask$, respectively.  
$\lossmse=\frac{1}{N}\sum_{i = 1}^{N}\lVert\img_i - \resimg_i\rVert_2^2$, 
where $N$ denotes the mini-batch size. $\losstask$ and $\lossrate$ were defined
earlier in this and the previous subsections.
\par\textbf{Loss weighting strategy:} 
Rather than using fixed loss weights, we propose a dynamic loss weighting 
strategy for an effective multitask training because: 
\textit{i)} The competing nature of the loss terms makes their respective 
gradients to worsen the performance of the others'. 
It is critical to have a right balance between the objectives in each update, 
which is very challenging for fixed loss weighting due to its inflexibility.
\textit{ii)} Exhaustive search for the optimal weights is very time-consuming 
\cite{uncertainty_loss_weighting}.

\begin{figure}
    \centering
    \centerline{\includegraphics[width=0.7\linewidth]
    {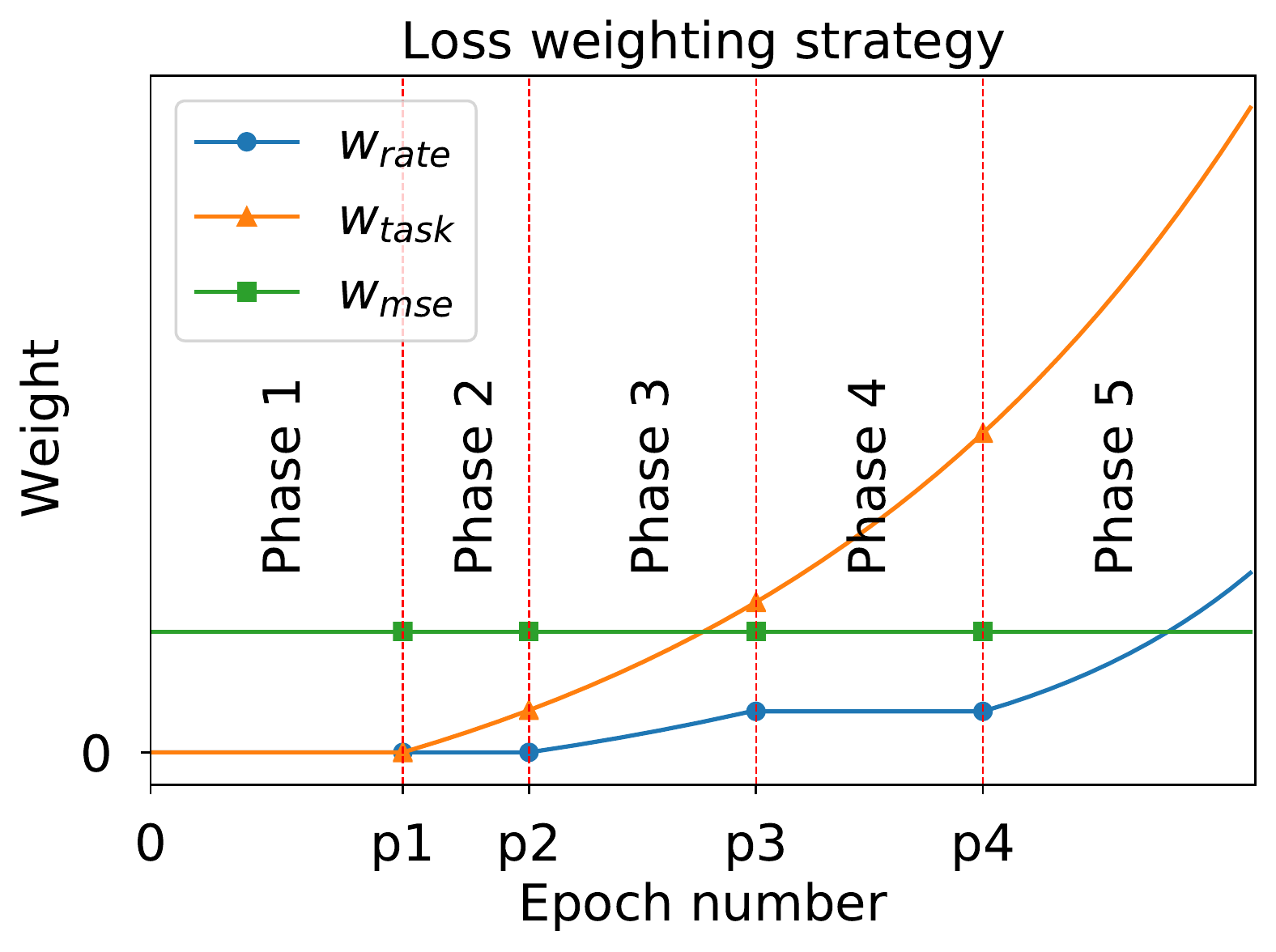}}
\caption{Loss weights evolution over iterations. There are 5 phases in this 
strategy, separated by vertical lines. Phase 1-5 respectively: base model 
training with $\lossmse$, introducing $\losstask$, introducing $\lossrate$, 
enhancing task performance, searching for optimal trade-offs when the system 
is stable.}
\label{fig:lossweight_strat}
\end{figure}

The task networks are trained on natural images, thus expect close-to-natural 
images as their input. 
In that light, we train a base model with only $\lossmse$ ($\wtask=\wrate=0,
\wmse=1$), 
which is capable of reconstructing images for a decent task performance. 
Then we fine-tune the base model by gradually raising $\wrate$ and $\wtask$ in 
different phases while keeping $\wmse=1$ as shown in 
\autoref{fig:lossweight_strat}, which eventually leads to the dominant impact 
of the gradients of $\lossrate$ and $\losstask$ on the accumulated gradients 
flow, effectively pushing the system to achieve an optimal task performance for 
a given bitrate constraint. 
Consequently, the same training instance is able to achieve a new 
rate-distortion performance for a different targeted bitrate after every 
iteration (see the results in \autoref{fig:rd_curve}). 
Learning rate decay is applied to keep the training stable. 
At inference time, a desired bitrate can be achieved by using the closest 
model's checkpoint (i.e., saved parameters) in terms of bitrate achieved 
during training on a validation set.
\ifdefined\ofincluded
    \newparagraph{Latent tensor fine-tuning:} At inference stage it is possible 
    to further optimize the system by adapting it to the content being encoded. 
    This way, even better rate-distortion performance can be achieved. 
    However, this content-adaptive optimization should be done only for the 
    components at encoder-side, otherwise adaptation-parameters need to be sent to 
    the decoder-side, resulting into bitrate overhead. 
    Instead of fine-tuning the encoder, it is sufficient to fine-tune 
    the latent tensor $\latent$, by back-propagating gradients of the loss with 
    respect to elements of $\latent$ through the frozen decoder and probability 
    model. 
    At inference stage, $\losstask$ is unknown to the encoder due to the fact that 
    the ground-truth for the tasks is not available. 
    However, the intermediate layer features of the task networks should be 
    correlated to those of other vision tasks to a certain degree. 
    Therefore, we fine-tune $\latent$ by using the gradients of $\Loss{inf} = 
    \Tilde{w}_{rate}\lossrate + \Tilde{w}_p\Loss{p}$, 
    where $\Tilde{w}_{rate}, \Tilde{w_{p}}$ denote the weights for the fine-tuning 
    loss terms and $\Loss{p}$ denotes a proxy feature-based loss. 
    In our experiments, we used a VGG-16 \cite{vgg} model pretrained on ImageNet 
    as our feature extractor and $\Loss{p} = \frac{1}{N}\sum_{i = 1}^{N}\left(\lVert 
    F_2(\img_i) - F_2(\resimg_i)\rVert_2^2 + \lVert F_4(\img_i) - F_4(\resimg_i)
    \rVert_2^2\right)$, 
    where N is the mini-batch size, $F_i(\Tensor{t})$ denotes the output of the 
    $i^{th}$ Max Pooling layer of the feature extractor given the input 
    $\Tensor{t}$. The weights are chosen to be $(\Tilde{w}_{rate}, 
    \Tilde{w_{p}})=(1, 0.1)$.
\fi

\section{Experiments and Discussion}
\label{sec:experiments}
\subsection{Experimental setup} The framework in \autoref{sec:proposed} is 
evaluated on two tasks: instance segmentation and object detection. 
The pre-trained models are provided by Torchvision\footnotemark, 
and the rest of the framework is implemented with Pytorch 1.5. 
We use the uncompressed dataset Cityscapes \cite{cityscapes} for training and 
testing our proposed models. 
Since the task models are pre-trained on COCO dataset \cite{coco},
we only evaluate results on the classes that are common between the two 
datasets: \textit{car, person, bicycle, bus, truck, train, motorcycle}.
\footnotetext{The pre-trained models can be found at 
\url{https://pytorch.org/docs/stable/torchvision/models.html}}
\newparagraph{Evaluation method and baseline:} We evaluated the performance of 
each compression method based on its average bitrate and task performance over 
500 images of the \textit{val} set. 
We use bits per pixel (BPP) as the bitrate metric and Mean Average Precision 
(mAP@[0.5:0.05:0.95])\cite{cityscapes} for the task performance.

As our baseline, we use the current state-of-the-art codec standard VVC
(reference software VTM-8.2 \cite{vtm82}, All-Intra configuration), under JVET common test 
conditions (CTC) \cite{ctc}. 
In order to achieve different bitrates, we encoded the \textit{val} set 
using 28 settings, which are the combinations of 7 quantization parameters 
(QP) (22, 27, 32, 37, 42, 47, 52) and 4 downsampling factors (100\% which 
corresponds to original resolution, 75\%, 50\% and 25\%). 
This results in 28 coded versions of the validation dataset. 
For the downsampled versions, data are upscaled to the original resolution 
before given to the task networks for task performance evaluation. 
\newparagraph{Loss weighting:} In these experiments, we model the loss weight 
values as functions of the epoch number. 
Concretely, the strategy shown in \autoref{fig:lossweight_strat} is formulated 
by
\begin{equation}
  \begin{split}
    \wmse&=1, \\
    \wtask&=\begin{cases} 
      0,& e < p_1\\
      4f_w(e-p_1,1.01),& e \ge p_1 
    \end{cases}, \\  
    \wrate&=\begin{cases} 
      0,& e < p_2\\
      2f_w(e-p_2,1.01),& p_2 \le e < p_3 \\
      c, &p_3 \le e < p_4 \\
      c+2f_w(e-p_4,1.02), &e \ge p_4 \\
    \end{cases},
  \end{split}
  \label{eq:lossstrat}
\end{equation}
where $p_1=50, p_2=75, p_3=120, p_4=165$, $f_w(x, a) = 10^{-3}(a^{x} - 1)$ and $c=2f_w(p_3-p_2-1,1.01)$.
\subsection{Experimental results}
\begin{figure}[ht]
    \begin{minipage}[b]{0.495\linewidth}
      \centering
      \centerline{\includegraphics[width=\linewidth]{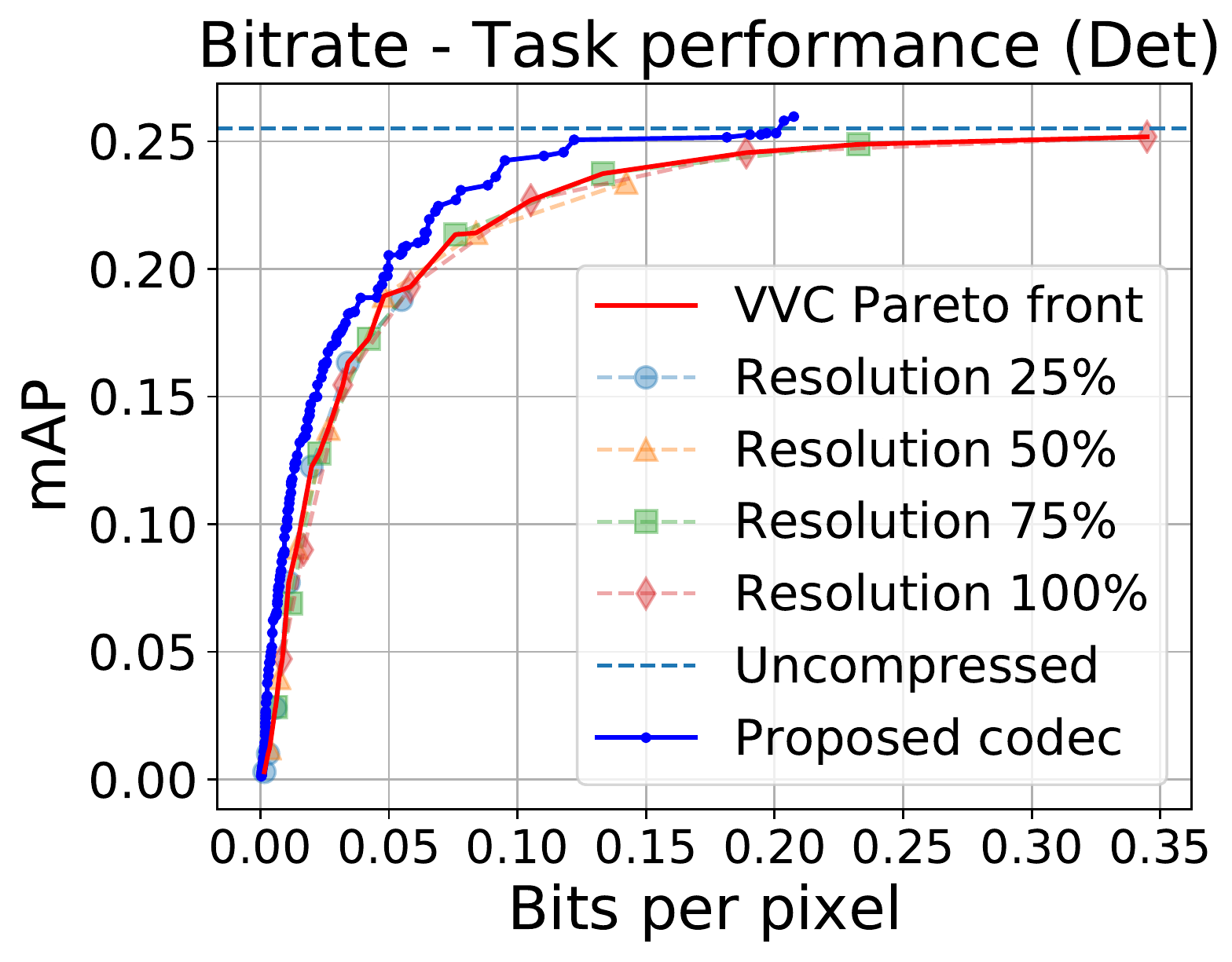}}
      \centerline{(a) Object detection}\medskip
      \label{subfig:det_curve}
    \end{minipage}
    \begin{minipage}[b]{0.495\linewidth}
      \centering
      \centerline{\includegraphics[width=\linewidth]{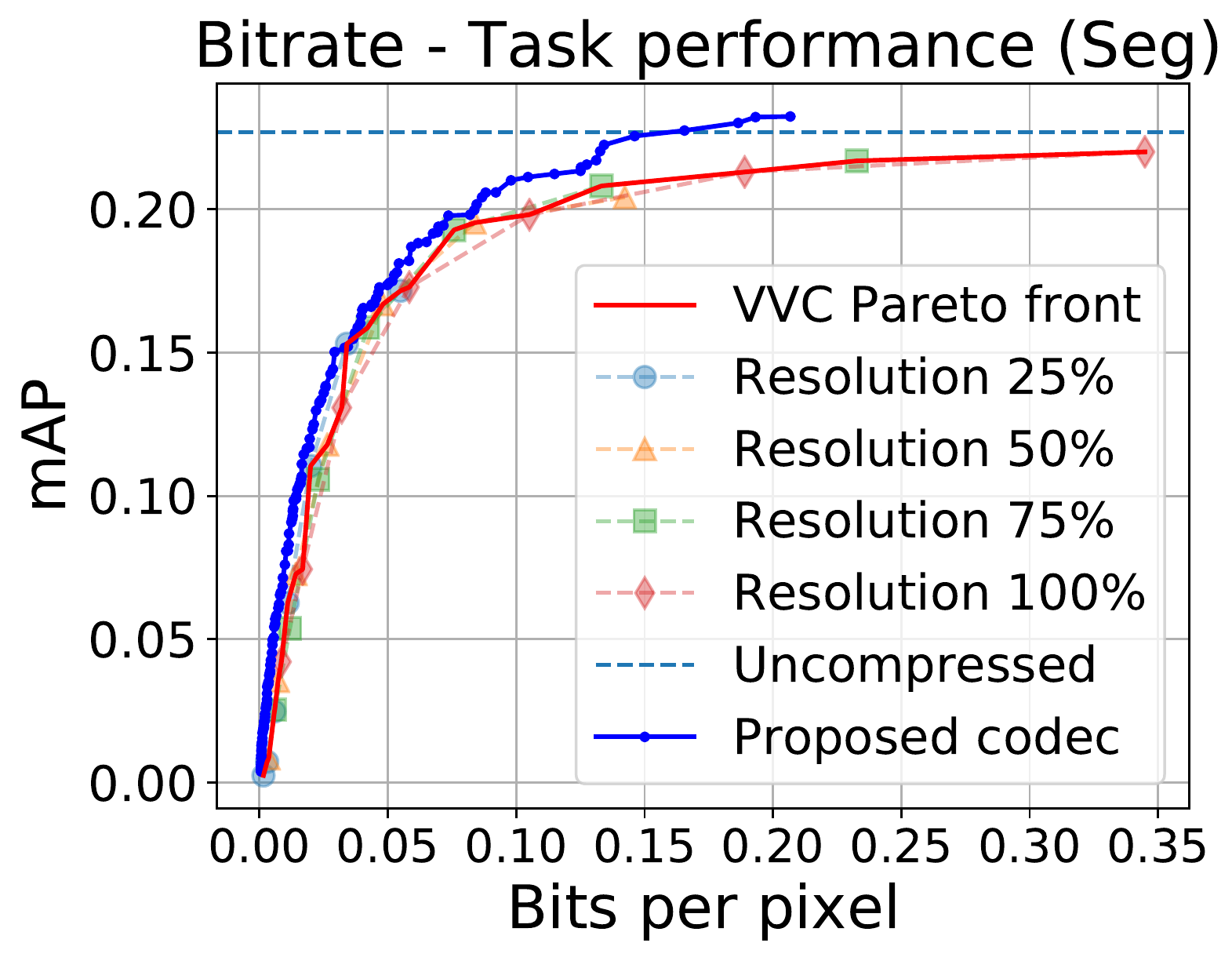}}
      \centerline{(b) Instance segmentation}\medskip
      \label{subfig:seg_curve}
    \end{minipage}

\caption{Rate-Performance curves. Each point on the dashed lines represents 
the performance of a VVC coded versions of the data with a certain QP and 
resolution combination.}
\label{fig:rd_curve}
\end{figure}

The Rate-Performance curves of our two systems against the corresponding baselines are 
shown in \autoref{fig:rd_curve}.
After every epoch of training set, we evaluated our models on \textit{val} set 
and obtained one data point for the curve ``Our method''. 
As the figure shows, our systems achieves superior performance for both tasks. 
Interestingly, both of the systems can achieve higher mAP than the best quality 
VVC settings (QP 22, 100\% resolution) while consuming only around 60\% of the 
bitrate. 
As shown in \autoref{tab:bd_rates}, our method on average saves 
\BDrateDet\%~bitrate for the same task performance level on object detection 
and \BDrateSeg\%~on instance segmentation, in comparison with the Pareto front of 
VVC anchors.
As a reference, although not directly comparable, for instance segmentation using 
the same task-NN architecture (Mask R-CNN) but trained on CityScapes 
\textit{train} set, \cite{vcm_feature_based} reports up to 9.95\% of bitrate 
saving for the following QPs: 12, 17, 22, 27, on the CityScapes \textit{val} set.
Our system contains only 1.5M parameters, and is also extremely fast: the average encoding time for a $2048 \times 
1024$ image in the \textit{val} set is around 0.15 seconds, with batch 
size of 1 on a single RTX 2080Ti GPU\footnotemark.
\footnotetext{As a reference, encoding a $2048 \times 1024$ image by VTM-8.2 software takes about 
125 seconds on a cluster with Intel Xeon Gold 6154 CPUs.}
\begin{table}[h]
  \small
  \centering
  \caption{Bjøntegaard Delta rate (BD-rate) with respect to task performance 
  against VVC anchors as in \autoref{fig:rd_curve}.}
  \begin{tabular}{lllllll}
    \hline
    && 100\%    & 75\%     & 50\%     & 25\%     & Pareto \\ \hline
    &Detection & -32.86 & -33.89 & -34.76 & -38.76 & -\BDrateDet \\ 
    &Segmentation & -33.77 & -28.58 & -29.98 & -28.04 & -\BDrateSeg \\ \hline
  \end{tabular}
  
  \label{tab:bd_rates}
\end{table}

\begin{figure}
    \begin{minipage}[b]{\outputfigwidth\linewidth}
      \centering
      \centerline{\includegraphics[width=\linewidth]
      {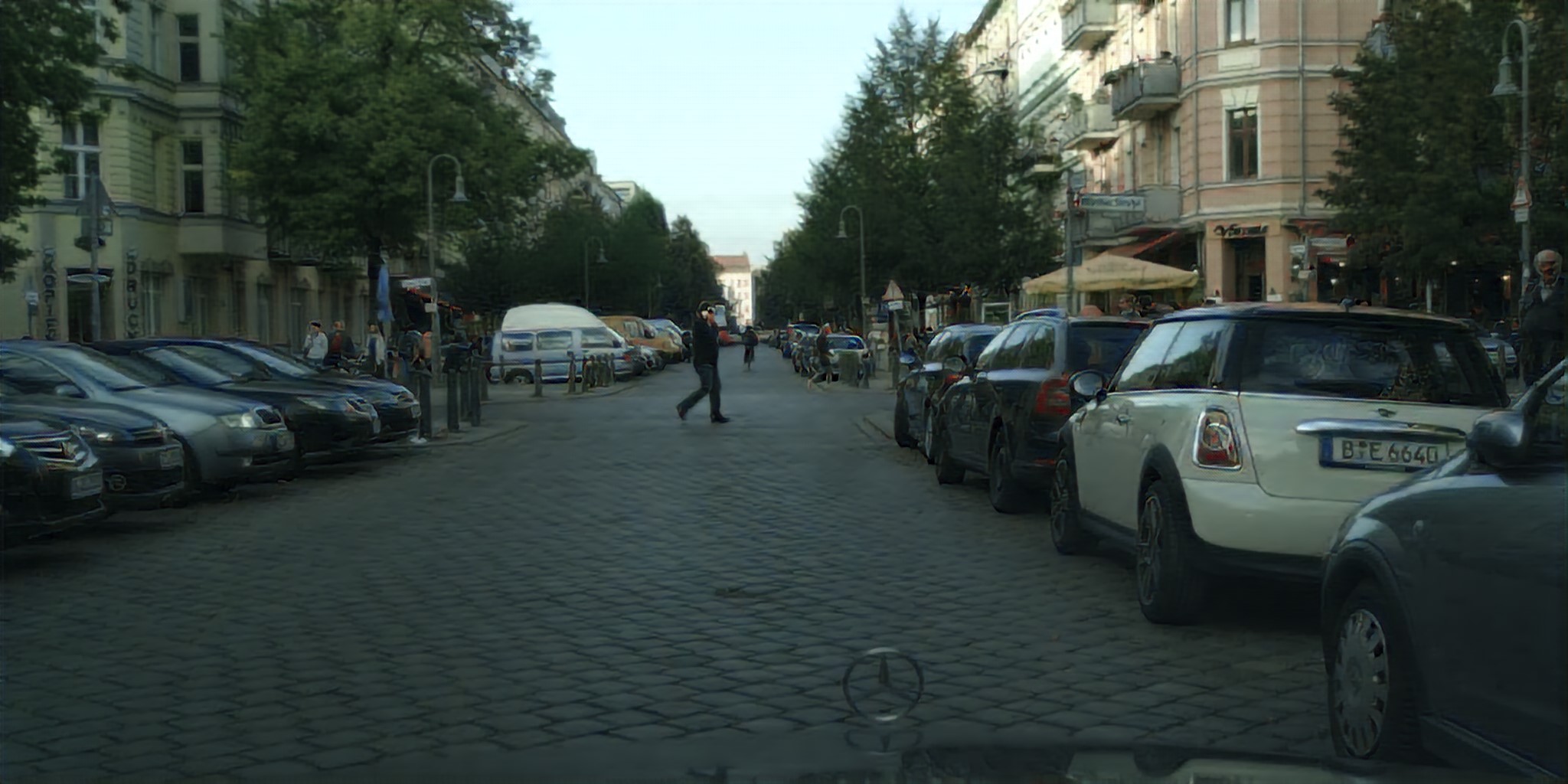}}
      \centerline{BPP: 0.247}\medskip
    \end{minipage}
    \begin{minipage}[b]{\outputfigwidth\linewidth}
      \centering
      \centerline{\includegraphics[width=\linewidth]
      {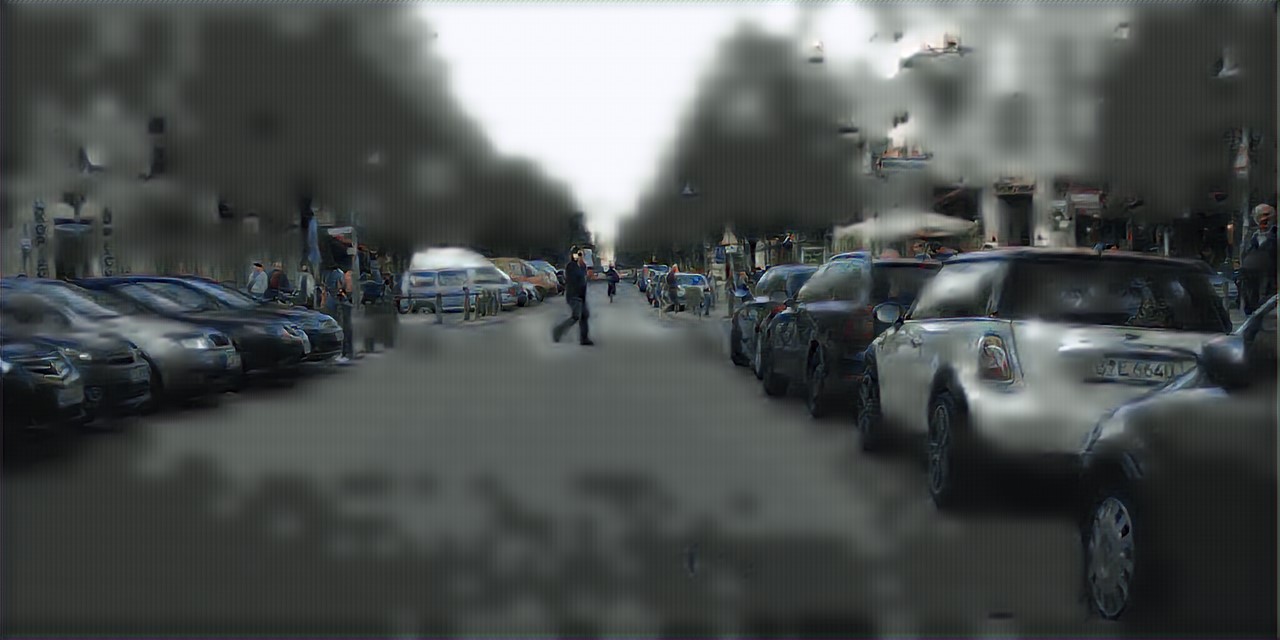}}
      \centerline{BPP: 0.045}\medskip
    \end{minipage}
    \begin{minipage}[b]{\outputfigwidth\linewidth}
      \centering
      \centerline{\includegraphics[width=\linewidth]
      {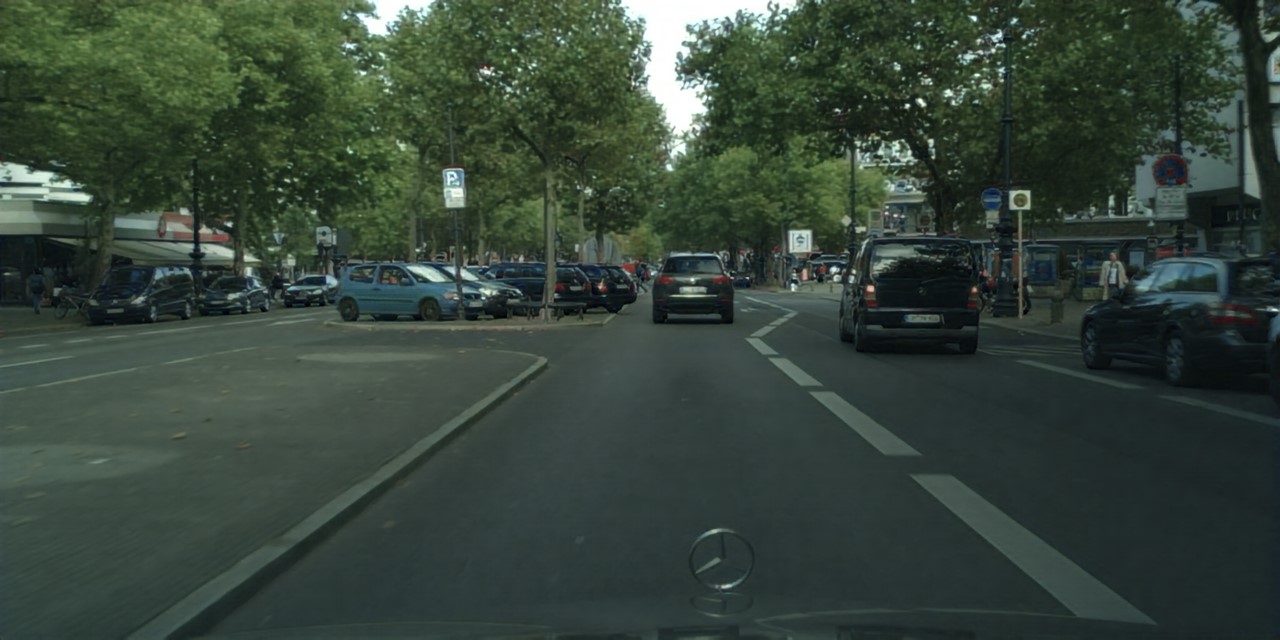}}
      \centerline{BPP: 0.241}\medskip
    \end{minipage}
    \begin{minipage}[b]{\outputfigwidth\linewidth}
      \centering
      \centerline{\includegraphics[width=\linewidth]
      {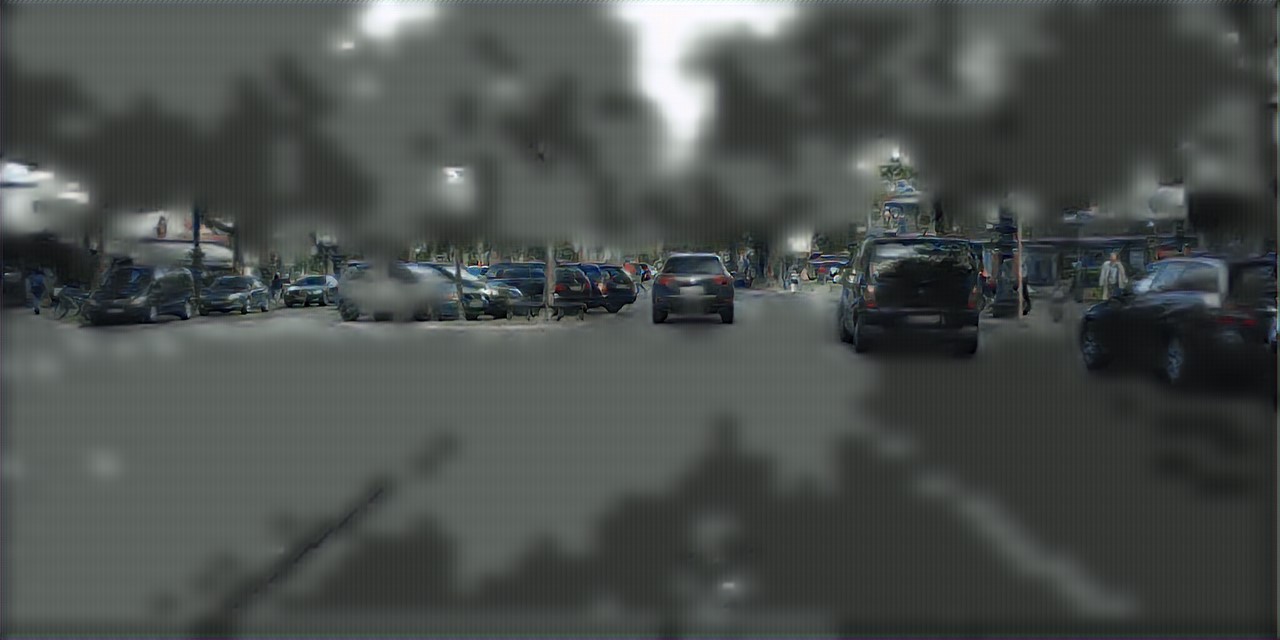}}
      \centerline{BPP: 0.030}\medskip
    \end{minipage}

\caption{Decoded outputs for instance segmentation in different bitrates. 
The background is fiercely suppressed in low bitrate.}
\label{fig:outputs}
\end{figure}

The reconstructed outputs in \autoref{fig:outputs} indicate that the system 
learns to compress more aggressively on the regions that are not too important 
to the task network. 
In particular, in low bitrate settings, the shapes and edges of the objects are 
preserved for the purpose of the machine tasks. 
Although the systems are trained on specific task network architectures, it 
preserves critical information from the input data that may be suitable for 
different tasks or different NN architectures, and this is a subject for our future work. 

\ifdefined\ofincluded
  \newparagraph{Latent fine-tuning:} As an incremental study, we have experimented 
  the latent fine-tuning technique on a similar system described in 
  \autoref{sec:proposed}, only with the entropy model replaced by a 
  super-resolution based model described in \cite{srec}. 
  In this experiment, we picked two random checkpoints: one at low and one at 
  high bitrate ranges and ran the fine-tuning for 30 iterations for every image 
  in the $val$ and reported the average measurements. 
  The results in \autoref{tab:of_results} reveal that this technique is helpful 
  for both bitrate reduction and task performance enhancement.
  \begin{table}
    \centering
    \caption{Results for latent tensor fine-tuning at inference stage}
    \begin{tabular}{l|llll}
      \hline
      \multirow{2}{*}{Rate range} & \multicolumn{2}{l}{Base model} & \multicolumn{2}{l}{Fine-tuned}  \\ \cline{2-5} 
                                  & BPP            & AP            & BPP            & AP             \\\hline
                                  
      Low                         & 0.054          & 0.162         & \textbf{0.052} & \textbf{0.162} \\
      High                        & 0.301          & 0.209         & \textbf{0.282} & \textbf{0.222} \\ \hline
    \end{tabular}
    
    \label{tab:of_results}
  \end{table}
\fi

\section{Conclusions}
\label{ch:summary}
We proposed and evaluated an efficient and fast end-to-end trained system for 
ICM. It shows that the proposed NN-based codec outperforms the state-of-the-art 
traditional codec VVC by a large margin when the compression is for 
machines. Additionally, we introduced a flexible loss weighting strategy 
for multi-task learning that both effectively trains the model towards the 
desired objectives and achieves optimal trade-off between them along the way at 
the same time. 
\ifdefined\ofincluded
    Finally, we suggested an inference stage fine-tuning technique 
    to further enhance the coding efficiency. We have begun exploring the 
    possibilities of adapting our system to multiple consumer tasks, making it a 
    robust alternative to the traditional standards for machine-consumption.
\fi





\bibliographystyle{ICASSP2021/IEEEbib}
\bibliography{refs}

\begin{thebibliography}{10}

\bibitem{vvc}
B.~Bross, J.~Chen, S.~Liu, and Y.-K. Wang,
\newblock ``{Versatile Video Coding} (draft 8),''
\newblock {\em {Joint Video Experts Team} (JVET), Document JVET-Q2001}, January
  2020.

\bibitem{selfdriving}
M.~Bojarski, D.~Del~Testa, D.~Dworakowski, B.~Firner, B.~Flepp, P.~Goyal, L.~D.
  Jackel, M.~Monfort, U.~Muller, J.~Zhang, X.~Zhang, J.~Zhao, and K.~Zieba,
\newblock ``End to end learning for self-driving cars,''
\newblock 2016.

\bibitem{vcm_feature_based}
K.~Fischer, F.~Brand, C.~Herglotz, and A.~Kaup,
\newblock ``Video coding for machines with feature-based rate-distortion
  optimization,''
\newblock {\em IEEE 22nd International Workshop on Multimedia Signal
  Processing}, p.~6, September 2020.

\bibitem{adapting_JPEGXS}
B.~Brummer and C.~de~Vleeschouwer,
\newblock ``Adapting {JPEG} {XS} gains and priorities to tasks and contents,''
\newblock in {\em 2020 {IEEE}/{CVF} Conference on Computer Vision and Pattern
  Recognition Workshops ({CVPRW})}. pp. 629--633, {IEEE}.

\bibitem{jointlearnedvvc}
Z.~Wang, R.-L. Liao, and Y.~Ye,
\newblock ``Joint learned and traditional video compression for p frame,''
\newblock in {\em 2020 {IEEE}/{CVF} Conference on Computer Vision and Pattern
  Recognition Workshops ({CVPRW})}. pp. 560--564, {IEEE}.

\bibitem{pixelCNN}
A.~V. Oord, N.~Kalchbrenner, and K.~Kavukcuoglu,
\newblock ``Pixel recurrent neural networks,''
\newblock New York, New York, USA, 20--22 Jun 2016, vol.~48 of {\em Proceedings
  of Machine Learning Research}, pp. 1747--1756, PMLR.

\bibitem{l3c}
F.~Mentzer, E.~Agustsson, M.~Tschannen, R.~Timofte, and L.~Van~Gool,
\newblock ``Practical full resolution learned lossless image compression,''
\newblock in {\em Proceedings of the IEEE Conference on Computer Vision and
  Pattern Recognition (CVPR)}, 2019.

\bibitem{scale_hyperprior}
J.~Ballé, D.~Minnen, S.~Singh, S.~J. Hwang, and N.~Johnston,
\newblock ``Variational image compression with a scale hyperprior,''
\newblock in {\em International Conference on Learning Representations}, 2018.

\bibitem{meanscale_hyperprior}
D.~Minnen, J.~Ballé, and G.~D. Toderici,
\newblock ``Joint autoregressive and hierarchical priors for learned image
  compression,''
\newblock in {\em Advances in Neural Information Processing Systems 31}, pp.
  10771--10780. Curran Associates, Inc., 2018.

\bibitem{lossweighting_overview}
T.~Gong, T.~Lee, C.~Stephenson, V.~Renduchintala, S.~Padhy, A.~Ndirango,
  G.~Keskin, and O.~H. Elibol,
\newblock ``A comparison of loss weighting strategies for multi task learning
  in deep neural networks,''
\newblock {\em {IEEE} Access}, vol. 7, pp. 141627--141632, 2019.

\bibitem{ans}
J.~Duda, K.~Tahboub, N.~J. Gadgil, and E.~J. Delp,
\newblock ``The use of asymmetric numeral systems as an accurate replacement
  for huffman coding,''
\newblock in {\em 2015 Picture Coding Symposium ({PCS})}, pp. 65--69.

\bibitem{relaxed_quantization}
J.~Ballé, V.~Laparra, and E.~P. Simoncelli,
\newblock ``End-to-end optimization of nonlinear transform codes for perceptual
  quality,''
\newblock in {\em 2016 Picture Coding Symposium ({PCS})}, pp. 1--5,
\newblock {ISSN}: 2472-7822.

\bibitem{fasterRCNN}
S.~Ren, K.~He, R.~Girshick, and J.~Sun,
\newblock ``Faster {R}-{CNN}: Towards real-time object detection with region
  proposal networks,''
\newblock in {\em Advances in Neural Information Processing Systems 28}, pp.
  91--99. Curran Associates, Inc., 2015.

\bibitem{maskRCNN}
K.~He, G.~Gkioxari, P.~Dollár, and R.~Girshick,
\newblock ``Mask {R}-{CNN},''
\newblock in {\em 2017 {IEEE} International Conference on Computer Vision
  ({ICCV})}, pp. 2980--2988,
\newblock {ISSN}: 2380-7504.

\bibitem{uncertainty_loss_weighting}
R.~Cipolla, Y.~Gal, and A.~Kendall,
\newblock ``Multi-task learning using uncertainty to weigh losses for scene
  geometry and semantics,''
\newblock in {\em 2018 {IEEE}/{CVF} Conference on Computer Vision and Pattern
  Recognition}, pp. 7482--7491,
\newblock {ISSN}: 2575-7075.

\bibitem{cityscapes}
M.~Cordts, M.~Omran, S.~Ramos, T.~Rehfeld, M.~Enzweiler, R.~Benenson,
  U.~Franke, S.~Roth, and B.~Schiele,
\newblock ``The cityscapes dataset for semantic urban scene understanding,''
\newblock in {\em 2016 {IEEE} Conference on Computer Vision and Pattern
  Recognition ({CVPR})}. pp. 3213--3223, {IEEE}.

\bibitem{coco}
T.-Y. Lin, M.~Maire, S.~Belongie, L.~Bourdev, R.~Girshick, J.~Hays, P.~Perona,
  D.~Ramanan, C.~L. Zitnick, and P.~Dollár,
\newblock ``Microsoft {COCO}: Common objects in context,''
\newblock 2015.

\bibitem{vtm82}
``Versatile video coding ({VVC}) reference software {VTM}-8.2,'' Available at:
  \url{https://vcgit.hhi.fraunhofer.de/jvetVVCSoftware_VTM} (Accessed on
  2020-10-20).

\bibitem{ctc}
F.~Brossen, J.~Boyce, K.~Suehring, X.~Li, and V.~Seregin,
\newblock ``{JVET} common test conditions and software reference configurations
  for sdr video,''
\newblock {\em Joint Video Experts Team (JVET), Document: JVET-N1010}, March
  2019.

\end{thebibliography}

\end{document}